\documentclass[sigconf]{acmart}

\copyrightyear{2020}
\acmYear{2020}
\setcopyright{rightsretained}
\acmConference[MLCAD '20] {2020 ACM/IEEE Workshop on Machine Learning for CAD}{November 16--20, 2020}{Virtual Event, Iceland}
\acmBooktitle{2020 ACM/IEEE Workshop on Machine Learning for CAD (MLCAD '20), November 16--20, 2020, Virtual Event, Iceland}
\acmPrice{15.00}
\acmDOI{10.1145/3380446.3430629}
\acmISBN{978-1-4503-7519-1/20/11}

\usepackage[ruled, linesnumbered]{algorithm2e}

\AtBeginDocument{%
  \providecommand\BibTeX{{%
    \normalfont B\kern-0.5em{\scshape i\kern-0.25em b}\kern-0.8em\TeX}}}

\begin{document}
\fancyhead{}

\title{Track-Assignment Detailed Routing Using Attention-based Policy Model With Supervision}

\author{Haiguang Liao}
\email{haiguanl@andrew.cmu.edu}

\affiliation{%
  \institution{Carnegie Mellon Unversity}
  \city{Pittsburgh}
  \state{PA}
  \postcode{15213}
}

\author{Qingyi Dong}
\email{qingyid@andrew.cmu.edu}

\affiliation{%
  \institution{Carnegie Mellon Unversity}
  \city{Pittsburgh}
  \state{PA}
  \postcode{15213}
}

\author{Weiyi Qi}
\email{weiyi@cadence.com}

\affiliation{%
  \institution{Cadence Design Systems}
  \city{San Jose}
  \state{CA}
  \postcode{95134}
}

\author{Elias Fallon}
\email{fallon@cadence.com}

\affiliation{%
  \institution{Cadence Design Systems}
  \city{San Jose}
  \state{CA}
  \postcode{95134}
}

\author{Levent Burak Kara}
\email{lkara@cmu.edu}

\affiliation{%
  \institution{Carnegie Mellon Unversity}
  \city{Pittsburgh}
  \state{PA}
  \postcode{15213}
}


\begin{abstract}
Detailed routing is one of the most critical steps in analog circuit design. Complete routing has become increasingly more challenging in advanced node analog circuits, making advances in efficient automatic routers ever more necessary. In this work, we propose a machine learning driven method for solving the track-assignment detailed routing problem for advanced node analog circuits. Our approach adopts an attention-based reinforcement learning (RL) policy model. Our main insight and advancement over this RL model is the use of \emph{supervision} as a way to leverage solutions generated by a conventional genetic algorithm (GA). For this, our approach minimizes the  Kullback-Leibler divergence loss between the output from the RL policy model and a solution distribution obtained from the genetic  solver. The key advantage of this approach is that the router can learn a policy in an offline setting with supervision, while improving the run-time performance nearly 100$\times$ over the genetic solver. Moreover, the quality of the solutions our approach produces matches well with those generated by GA. We show that especially for complex problems, our supervised RL method provides good  quality solution similar to  conventional attention-based RL without comprising run time performance. The ability to learn from example designs and train the router to get similar solutions with orders of magnitude run-time improvement can impact the design flow dramatically, potentially enabling increased design exploration and routability-driven placement.
\end{abstract}




\maketitle

\section{Introduction}
\begin{figure}[htbp]
    \centering
    \includegraphics[width=0.38\textwidth]{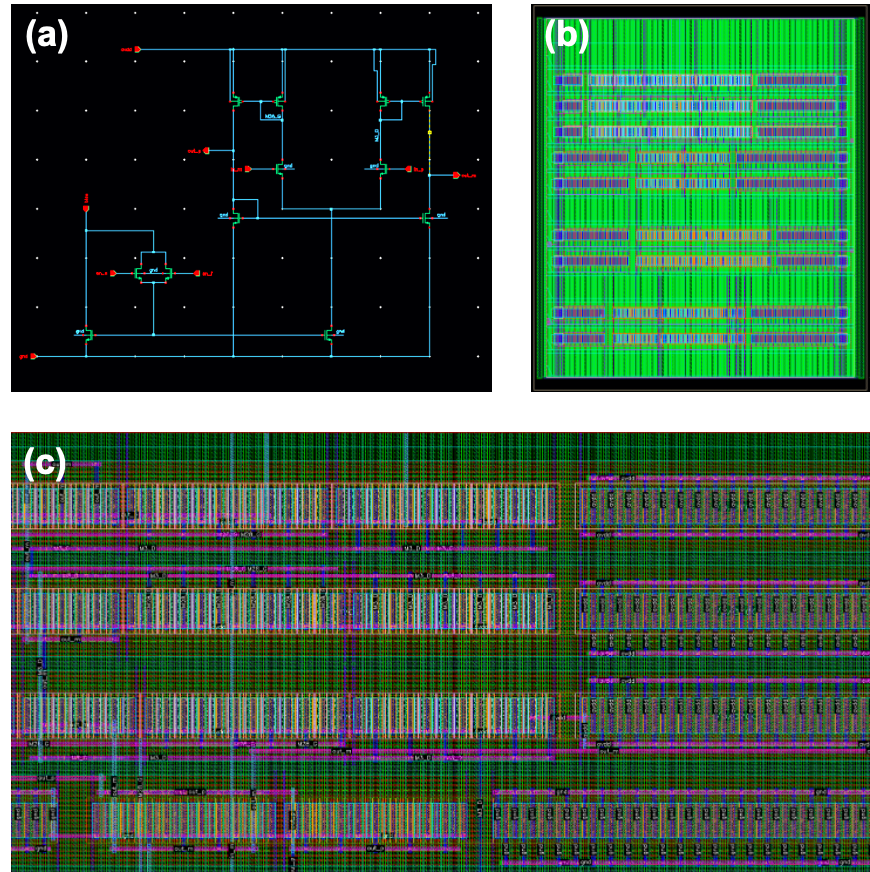}
    \caption{Advanced node technologies analog circuits \textit{biasamp} (random generated) (a) schematic and (b) design solutions given by Cadence Virtuoso, (c) magnified version of design solutions given by Cadence Virtuoso.}
    \label{Virtuoso}
\end{figure}

Detailed routing is a critical step in the physical design of VLSI analog circuits, where actual routes are constructed based on placement and global routing steps.  In advanced node technologies, the increased complexity of the  problems renders conventional routers difficult to scale to and performs efficiently on the types of problems shown in Fig.~\ref{Virtuoso}.  While prior analog routing works \cite{malavasi1993area,lampaert1996analog,zhu2019geniusroute} have proposed template-based, simulation-based, and heuristic-based methods, the key issue of a generalizable net sequencing and pin-pair connection remains largely unsolved.  Moreover, a lack of suitable mechanisms that can learn from past routing solutions prevents the use of such valuable data toward an efficient solution of new routing problems. Based on these observations, new data-driven methods have been recently  proposed for analog routing and associated classes of problems, \cite{liao2020attention,mangalagiri2019analog,mirhoseini2020chip}. Most closely related to our work, Liao \textit{et al.} propose an attention-based  reinforcement learning (ARL) method for detailed routing \cite{liao2020attention}. However, the approach also suffers from at-times unstable training  and fails to take advantage of the already computed genetic routing solutions, as no labeled data can be used to supplement the ARL model. 

\subsection{Our Contributions}

In this work, we propose a new analog router model that builds on ARL, whose training can now be additionally supervised with past solutions. We refer to our routing method as supervised reinforcement learning (SRL). Once trained, both ARL and the new SRL  provide nearly 100$\times$ acceleration in runtime performance over the GA while generating solutions that are comparable in quality to those generated by GA. Moreover, we show that as the problem complexity increases, SRL solution quality remains more similar to that of GA on previously unseen problems.  

Our work thus demonstrates through the use of supervised RL, routing solutions can be generated significantly faster over conventional methods without compromising solution quality.  This capability enables significant future potential flow innovations. With a learn-by-example capability in routing, the user could provide examples of manufacturing proven designs which were routed by experts, and have the router learn the key decisions necessary to match those results. The router then has the potential to implicitly learn the user's underlying quality factors. Subsequently being able to create new routing based on the learned policy, at significantly faster run-time than traditional routers, enables integration of those results into other steps of the design creation flow. For instance, placement algorithms  can utilize fast routers in their inner loop as objective functions in their optimization.

\section{PRELIMINARIES}
\subsection{Track-assignment detailed routing}
Routing  is typically addressed in two sequential steps: \textit{global routing} and \textit{detailed routing}. Global routing approximately allocates routing resources into sub-regions while detailed routing constructs the actual routes in the individual sub-regions while satisfying various design rule constraints. In this work, we focus on routing completion success and overall wirelength as the two primary considerations of a detailed router's solution quality. We thus abstract  any applicable design rules into Width Spacing Patterns (WSP), where track patterns consisting of various width and spacing parameters for metal wires are prescribed. By restricting the routes on the WSP rows and tracks, many design rules associated with full custom designs can be imposed a priori using a modified weighted bipartite matching formulation \cite{liao2020attention}. 

The use of WSP for design rule abstraction, net decomposition into pin-pairs, and the use of a pattern router for eventual routing allows all three methods (GA, ARL, and the new SRL) discussed in this work to be compared on a common basis. Each of these methods aims to produce a sequencing of the pin-pairs such that when the pairs are routed in that sequence using the pattern router, the overall quality  of the solution (measured in terms of the number of completed routes and total wirelength)  can be maximized. 

\subsection{Attention-based reinforcement learning policy model}
Recent studies have demonstrated  the success of reinforcement learning for combinatorial optimization  problems with better generalization capabilities and runtime performance  over heuristic based or exact methods \cite{bello2016neural,barrett2019exploratory,li2018combinatorial}. Among these works, Kool \textit{et al.}  \cite{kool2018attention} uses an attention-based reinforcement learning policy model to solve the Traveling Salesman Problem  problem (TSP). They achieve state-of-the-art performance with an ability to generalize to unseen problems over previous approaches \cite{vinyals2015pointer, nazari2018reinforcement}. 
In solving the TSP with the policy model, a solution can be defined as a tour $\pi$, which consists of an ordered sequence of nodes (cities). At each step, the policy model generates a probability distribution $p_{\theta}(\pi_t|s, \mathbf{\pi}_{1:t-1})$ over the possible nodes to visit next. A solution for a TSP problem can be obtained in $n$ steps, where $n$ represents the number of cities in the TSP problem. Based on this mechanism, the work defines a problem policy by multiplying the probability distributions at each step, as shown in Eqn.~\ref{problem_policy}.

\begin{equation}
\label{problem_policy}
    p_{\theta}(\mathbf{\pi}|s) = \prod_{i=1}^n p_{\theta}(\pi_t|s,\mathbf{\pi}_{1:t-1})
\end{equation}

In order to obtain a robust policy model, the policy model needs to be optimized  with respect to its parameters. Based on the problem policy and the policy gradient theorem used in REINFORCE, the gradient of the loss function for the policy model can be stated as: $\nabla_{\theta} E_{\pi\sim p_{\theta}(\pi|s)}[L(\pi)]. $ This can be calculated in  Eqn.~\ref{pgt} based on policy gradient theorem \cite{sutton2018reinforcement}, which provides a feasible way to calculate the otherwise intractable gradient information to optimize the model parameters:

\begin{equation}
\label{pgt}
    \nabla_{\theta} E_{\pi\sim p_{\theta}(\pi|s)}[L(\pi)] = E_{\pi\sim p_{\theta}(\pi|s)} [L(\pi)\nabla_{\theta}log p_{\theta}(\pi|s)]
\end{equation}

\begin{figure}[thpb]
    \includegraphics[width=0.45\textwidth]{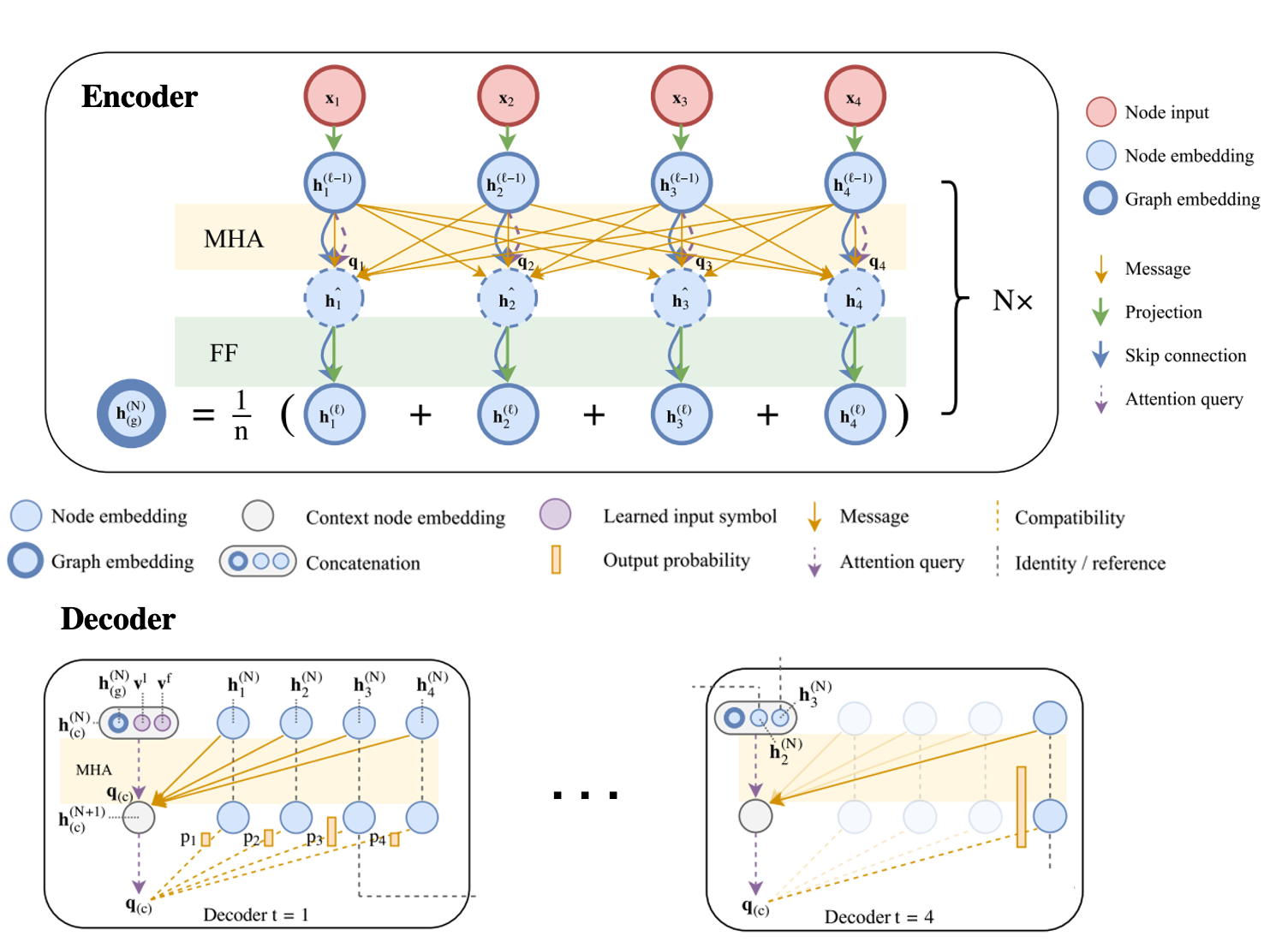}
    \caption{Schematic plot of the attention-based model. Adopted from \cite{kool2018attention}}
    \label{EncodeDecoder}
\end{figure}

The structure of the applied policy model  is an attention-based encoder-decoder (Fig.~\ref{EncodeDecoder}). This model is as a variation of Graph Attention Networks whose advantage is the ability to solve a sequential problem in a sequence-independent way.

Driven by the similarity between the analog detailed routing problem and the TSP, previous work \cite{liao2020attention} has successfully applied an attention-based reinforcement learning (ARL) policy model to solve the track-assignment detailed routing problems. In that work, ARL is able to solve detailed routing problems in a generalizable way. The most significant contribution of ARL is a new, fast method for sequencing the pin pairs extracted from the netlists. Without an appropriate sequence, detailed routing often needs to be solved using computationally demanding and sub-optimum rip-up and re-route strategies \cite{dees1982automated}. The ARL is trained without any supervision. This works builds upon ARL and introduces supervision provided by the GA solutions. We call the resulting approach  supervised RL (SRL) for routing and discuss it next. This work is inspired by recent works in complementing RL with supervision \cite{shelhamer2016loss,zhang2018applying}

\subsection{Policy model trained with supervision}
In order to enable supervision for the policy model, a target (label) similar to the output of the problem policy must be generated. We use the GA solution as labeled data. We transform these solutions obtained in the populations to probability distributions for use in the policy SRL model.  Specifically, consider a given problem instance $s$ which consists of $n$ two-pin pairs to be ordered and routed using a pattern router. After GA solver is applied, a set of optimized (but not necessarily optimal) sequences $\{\mathbf{\pi}_1, \mathbf{\pi}_2, ..., \mathbf{\pi}_m \}$ of the $n$ two-pin pairs can be obtained. At each location $q_t, t=1,...,n$ of the sequence, an empirical distribution among two-pin pairs can be obtained. Based on this  distribution, by multiplying the $q_t, t=1,...,n$, a label distribution for the policy of problem instance $s$ can be obtained. For ease of algebraic manipulation, we utilize the log-label in the form:

\begin{equation}
\label{q}
    log (q(s)) = log (\prod_{t=1}^n q_t) = \sum_{t=1}^n log (q_t)
\end{equation}

Both $q_t$ and $q(s)$ are vectors and the log operations applied to the individual components in the vector, thus $log (q(s))$ is also a vector. With this label and problem policy given by the attention-based policy model, a loss is formulated based on Kullback-Leibler (KL) divergence:

\begin{equation}
\label{kl}
    L(s) = D_{KL}(p_{\theta(s)} |q(s) ) = p_{\theta(s)}^T  (log(p_{\theta(s)})-log(q(s)))
\end{equation}

This loss encodes the cross-entropy loss between two distributions; one from the policy model and the other one from the GA solutions. We use a discretized version:

\begin{equation}
\label{kl_dis}
    L(s) = \sum_{i=1}^n p_{\theta(s)}(i) (log(p_{\theta(s)}(i))-log(q(s))(i))
\end{equation}

\subsection{Problem Formulation}
\begin{figure*}[thpb]
\centering
\includegraphics[width=0.7\textwidth]{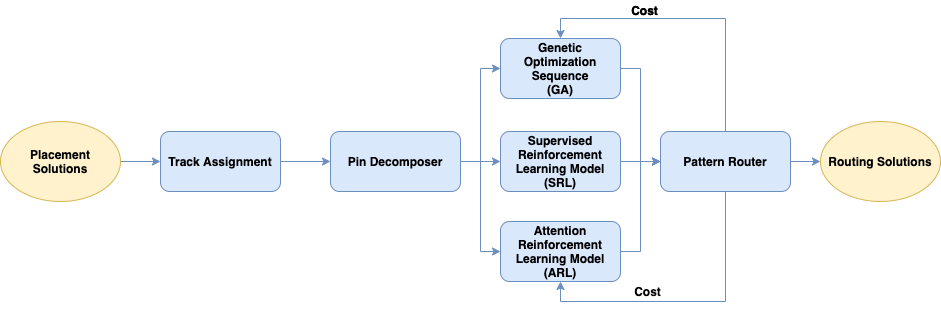}
\captionof{figure}{The flow of proposed learning-based solutions for track-assignment detailed routing.}
\label{flow}
\end{figure*}

Fig.~\ref{flow} shows the proposed learning-based solution flow for track-assignment detailed routing. 
The input of the flow is a set of placement solutions $\mathbf{S}=\{s_1, s_2, ..., s_m\}$ for a sub-region of a given analog circuit design to be routed. The placement solutions (of a sub-region of a chip) specify the locations of the devices and netlists containing the grouping information of the devices that need to be connected in the detailed routing stage. It is assumed that the placement solution sets $\mathbf{S}$ for a specific region of a given analog design follows a certain  distribution (which is the probability of different devices  appearing at the different locations in the design space), instead of a random distribution. This is based on the observation that placement solutions  are typically generated by considering similar objectives such as device density or  potential drops. After the track assignment and pin decomposition, a detailed routing for each placement $s_i$ is formulated as the connection of a set of two-pin pairs $N = \{t_1, t_2, ..., ,t_n\}$ with predetermined spatial locations and following a  sequence $\mathbf{\pi} = \{\pi_1,\pi_2, ..., \pi_n\}$. The number of possible sequences  is governed by all the permutations within the two-pin pairs set $N$. The sequential detailed routing problem is formally formulated as follows:

\textbf{Sequential Detailed Routing}: Consider a set of two-pin pairs $N = \{t_1, t_2, ..., ,t_n\}$ on a detailed routing region represented as a graph $G(V,E)$, in which pins are distributed on different vertices and edges. Only horizontal and vertical directions are the feasible routing directions, with a routing capacity of 1. The sequential detailed routing engine solves  the detailed routing problem in two steps. First, an optimized sequence $\mathbf{\pi}$ of two-pin pairs are generated by a solver. Then, following the sequence $\mathbf{\pi}$, a pattern router routes individual two-pin pairs  sequentially while sharing the same graph $G(V,E)$. 

\textbf{Objective Function}: The metrics to be optimized include: (1) the total wirelength of all two-pin pair routes, (2) the total number of openings described as the unrouted two-pin pairs. When a  two-pin pair can not be routed, the total number of openings increased by one. For ease of implementation, a weighted sum of the total number of openings and total wirelength is used as the objective function, hence cost, to minimize:

\begin{equation}
\label{costeqn}
    Cost(s) = w_1 * Wirelength + w2 * \#Open
\end{equation}

The lower the cost, the better the solution quality is. The weights of the two terms can be adjusted by the end user to bias the two components of the objective differently. 

\section{ALGORITHMS}
As shown in Fig.~\ref{flow}, we analyze the GA, ARL, and SRL methods. All three methods use the same the track assignment method and pin decomposition algorithm (Kruskal’s algorithm \cite{kruskal1956shortest}). All three methods use the same pattern router consisting of ``L'' and ``Z'' which provides a common basis for comparing the algorithms. The GA-based sequencing optimizes the sequence  $\mathbf{\pi}$ for a given problem through multiple generations of  iterations. It uses permutation-based crossover and mutation operations for creating off-springs. The ARL is used as a  way to learn to generate optimized sequences through self-training on a set of problems in previous work, which is built upon the attention-based reinforcement learning policy model.

In this work, we propose a supervised version of ARL, we call SRL, based on the KL-divergence loss and the GA solutions as supervision, as shown in Algorithm~\ref{supervised_algo}. During  training, $60\%$ of placement solutions of a given design is used as training data, $20\%$ is used for validation and $20\%$ is used as testing. Given the training problems, GA first generates results in the form of a distribution $q$ based on formulation Eqn.~\ref{q} after 10 generations. After  $q$ is obtained from the GA, SRL training is initiated by calculating the KL-divergence loss between $q$ distribution and distribution output $p_{\theta}$ from the policy model. The loss is then used for updating the policy model parameters by backpropagation, which we perform for 100 epochs.   

Once training  is complete, SRL provides pin pair sequences for previously unseen problems (placements) from the same data set in a forward fashion. The runtime computational cost of the trained SRL (as well as ARL) is negligible compared to GA, as only matrix multiplications are involved during run time.

\begin{algorithm}
\caption{SRL model.}
\label{supervised_algo}
\SetAlgoLined
\SetKwInOut{Input}{Input}\SetKwInOut{Output}{Output}
\Input{Number of epochs E, batch size B, training set T (with GA-based label, $q$)}
\Output{Sequence based on best policy}
Init $\theta$ $\leftarrow$ $\theta$ \;
\For{epoch=1,...,$E$}{
 \For{batch=1,...,$B$}{
    $t_i$ $\leftarrow$ SampleInstance() $\forall i \in {1,...,T}$\;
    $\pi_i$ $\leftarrow$ SampleRollout($t_i$, $p_{\theta}$) $\forall i \in {1,...,T}$\; 
    
    $L$ $\leftarrow$ $\sum_{i=1}^B{p^T_{\theta}(i)(log(p_{\theta}(i))-log(q(i)))}$\;
    $\theta$ $\leftarrow$ Adam($\theta$, $\Delta L$)\;
 }
}
\end{algorithm}

\section{EXPERIMENTAL RESULTS}

\textbf{Data:} To evaluate the three methods and compare their performance, we use two kinds of analog design problems as follows:

\begin{enumerate}
    \item \textit{biasamp}: This consists of comparators and OpAmps, and less than 100 devices in each potential design. The \textit{biasamp} problems are randomly generated placement solutions to explore the model properties and therefore does not represent feasible designs for actual analog circuits. 
    
      \item \textit{sarfsm}: This is a subcircuit of an Analog-to-Digital Converter, with hundreds of devices in each  potential design. This problem is more complex compared to \textit{biasamp}.
\end{enumerate}

Both problems  are representative of advanced node technologies (sub-16 nm technology) analog circuits design problems. In order to analyze the sample efficiency for training the SRL models, we  generate three data sets consisting of a different number of device configurations (designs):

\begin{enumerate}
    \item $biasamp500$: biasamp with 500 device configurations
    
     \item $biasamp5000$: biasamp with 5000 device configurations
     
     \item $sarfsm500$: sarfsm with 500 device configurations
    
\end{enumerate}

\textbf{Implementation:} All three routing algorithms and the whole flow is implemented in \textbf{Python3.6} with the machine learning framework \textbf{Pytorch}. Experiments are run on a workstation with an Intel Core i7-6850 CPU without GPU acceleration. All models are trained using the following  parameters: training epochs: 100, batch size: 5, optimizer: Adam. 

\textbf{Training Time:} Training time varies with problem size and the size of the data set.  SRL  takes 6 minutes for training in $biasamp500$, and 30 minutes for training in $sarfsm500$. ARL's training time is virtually identical to that of SRL's. While currently not used, the MHA mechanism of both the SRL and ARL can be readily parallelized, which is the subject of our future work.   

\begin{figure}[htbp]
\centering
\includegraphics[width=0.47\textwidth]{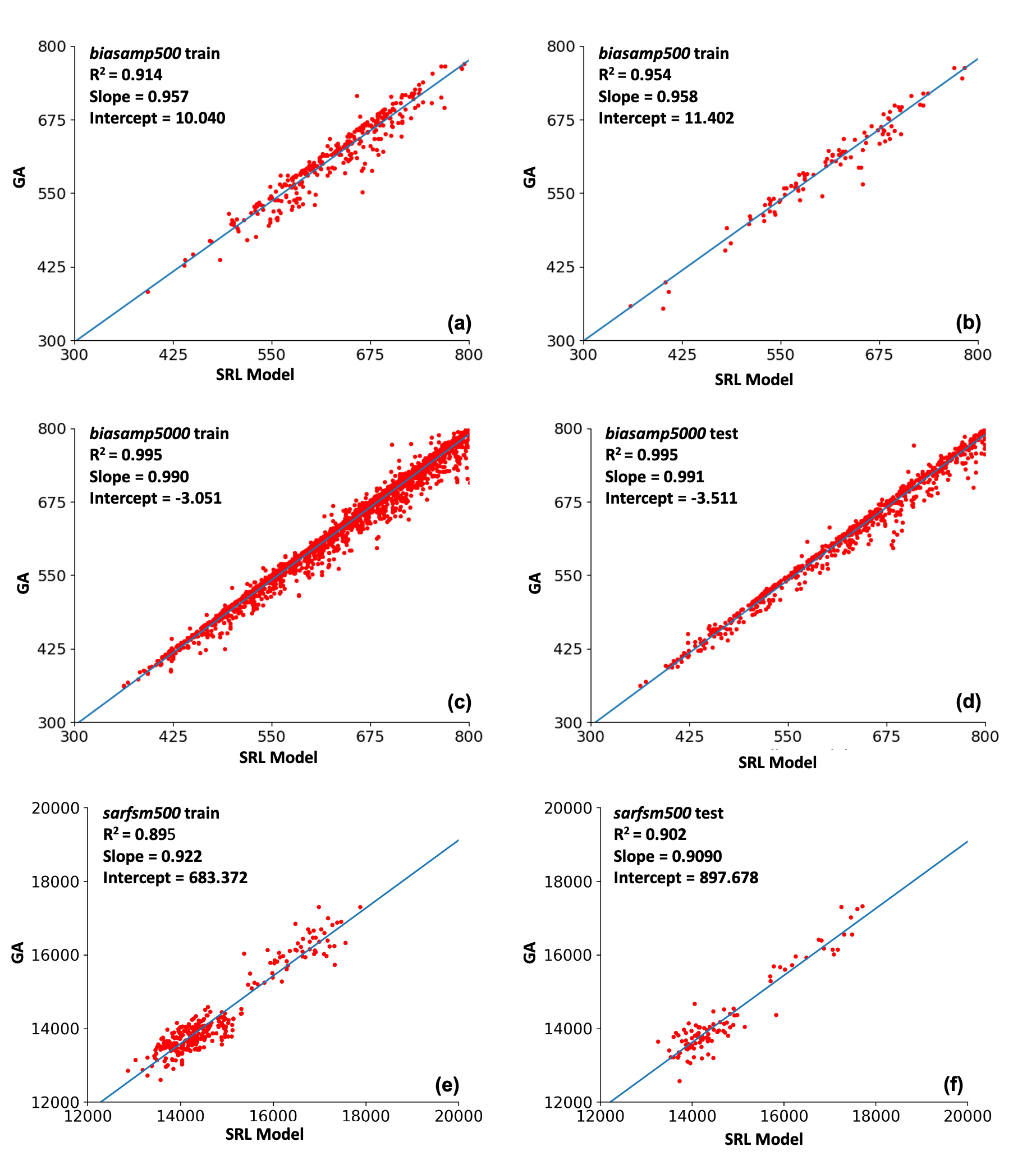}
\caption{Correlation analysis between SRL solutions' cost and GA solutions' cost on (a) \textit{biasamp500} training set, (b) \textit{biasamp500} test set, (c) \textit{biasamp5000} training set,  (d) \textit{biasamp5000} test set, (e) \textit{sarfsm500} training set, (f) \textit{sarfsm500} test set.}
\label{super_corr}
\end{figure}

\subsection{SRL performance}
Fig.~\ref{super_corr} compares the results of SRL against the GA solutions with respect to the solution cost, with correlation analysis between these two algorithms. Each red point corresponds to a unique device configuration problem that is routed both by SRL and GA. In the three data sets studied (rows), both the training and test sets (columns) reveal a strong linear correlation between the SRL and GA solutions. The resulting slopes (<1.0) and insignificant intercepts (compared to the range of cost) indicate that SRL solution quality is slightly less than that of GA's. This is further quantified in the next few paragraphs. 

For the same data set, the correlation strength between SRL  and GA  remain consistent across the training and test sets. This indicates that the supervised training on  can effectively generalize  to the previously unseen test sets. The second observation is that increasing the size of data set can effectively increase the SRL's strength of correlation with GA in both the training and test set, as seen from \textit{biasamp500} (Fig.~\ref{super_corr}a and Fig.~\ref{super_corr}b) and \textit{biasamp5000} (Fig.~\ref{super_corr}c and Fig.~\ref{super_corr}d).  Specifically, after significantly increasing the number of placement solutions in the data set, the $R^2$ score between SRL  and GA increase from 0.954 to 0.995. 

\begin{figure}[htbp]
\centering
\includegraphics[width=0.49\textwidth]{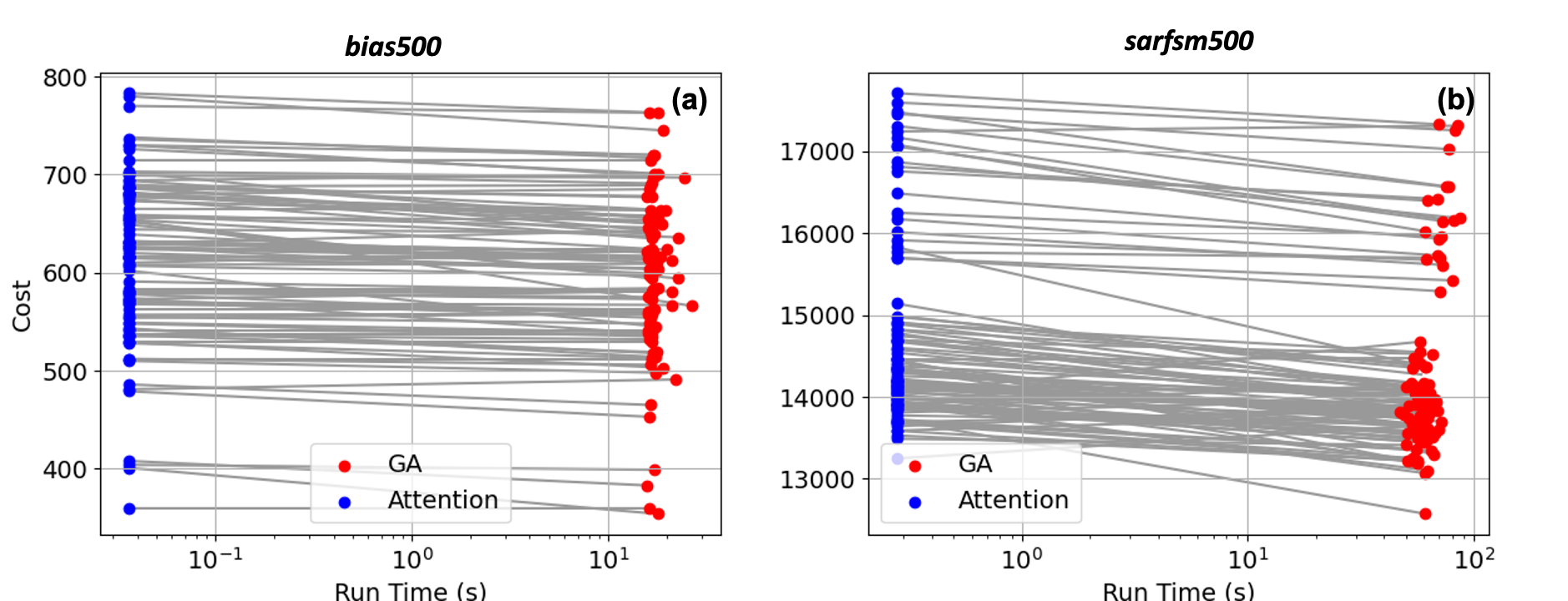}
\caption{Cost vs. Runtime comparison between SRL and GA for test set of (a) \textit{biasamp500} and (b) \textit{sarfsm500}.}
\label{cost_runtime}
\end{figure}
\begin{figure}[htbp]
\centering
\includegraphics[width=0.47\textwidth]{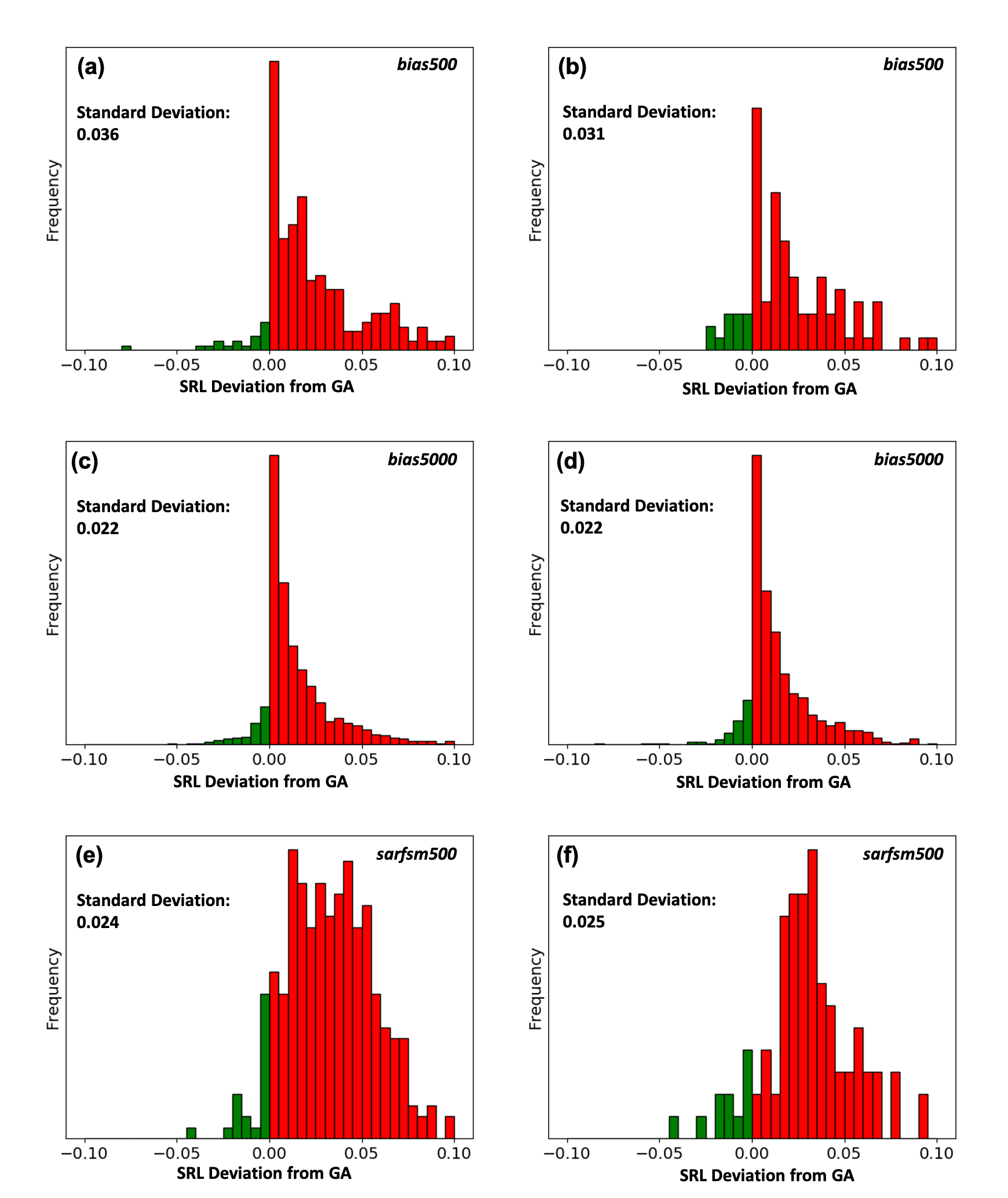}
\caption{Histograms showing deviation of SRL from  GA for training sets of (a) \textit{biasamp500}, (b) \textit{biasamp5000}, (c) \textit{sarfsm500}, and test set of (d) \textit{biasamp500}, (e) \textit{biasamp5000}, (f) \textit{sarfsm500}.}
\label{hist}
\end{figure}

The cost vs. run time comparisons of the SRL and GA models on test sets \textit{bias500} and \textit{sarfsm500} are shown in Fig.~\ref{cost_runtime}. In the plots, blue dots correspond to the SRL cost and run time, and red dots correspond to GA's cost and run time. The red-blue point pairs connected with gray lines correspond to the same problem.  As seen, the run time performance or SRL is two orders of magnitude better than that of the GA's, while only encountering a slight degradation in the solution quality. To show the difference in the quality of the solutions between SRL and GA, Fig.~\ref{hist} shows the histograms of signed deviations of SRL  model's cost from the corresponding GA-based results on different data sets. In these plots, the random variable is $\frac{Cost^{SRL} - Cost^{GA}}{Cost^{GA}}$. By comparing the results on different data sets, it is seen that larger data sets  and more complex designs result in a wider variation in the  SRL solution relative to GA. In all the data sets, SRL's cost are within $[-8\%,10\%]$ of the GA  cost, while in most cases SRL's  results are within $\pm 5\%$  of the GA results. This demonstrates  that for the data sets studied in this work,  the SRL policy would be at most $10\%$ worse than GS, while generating solutions around $1\%$ of the time it takes for GA.

\subsection{SRL Comparison with ARL}
\begin{figure}[htbp]
\centering
\includegraphics[width=0.45\textwidth]{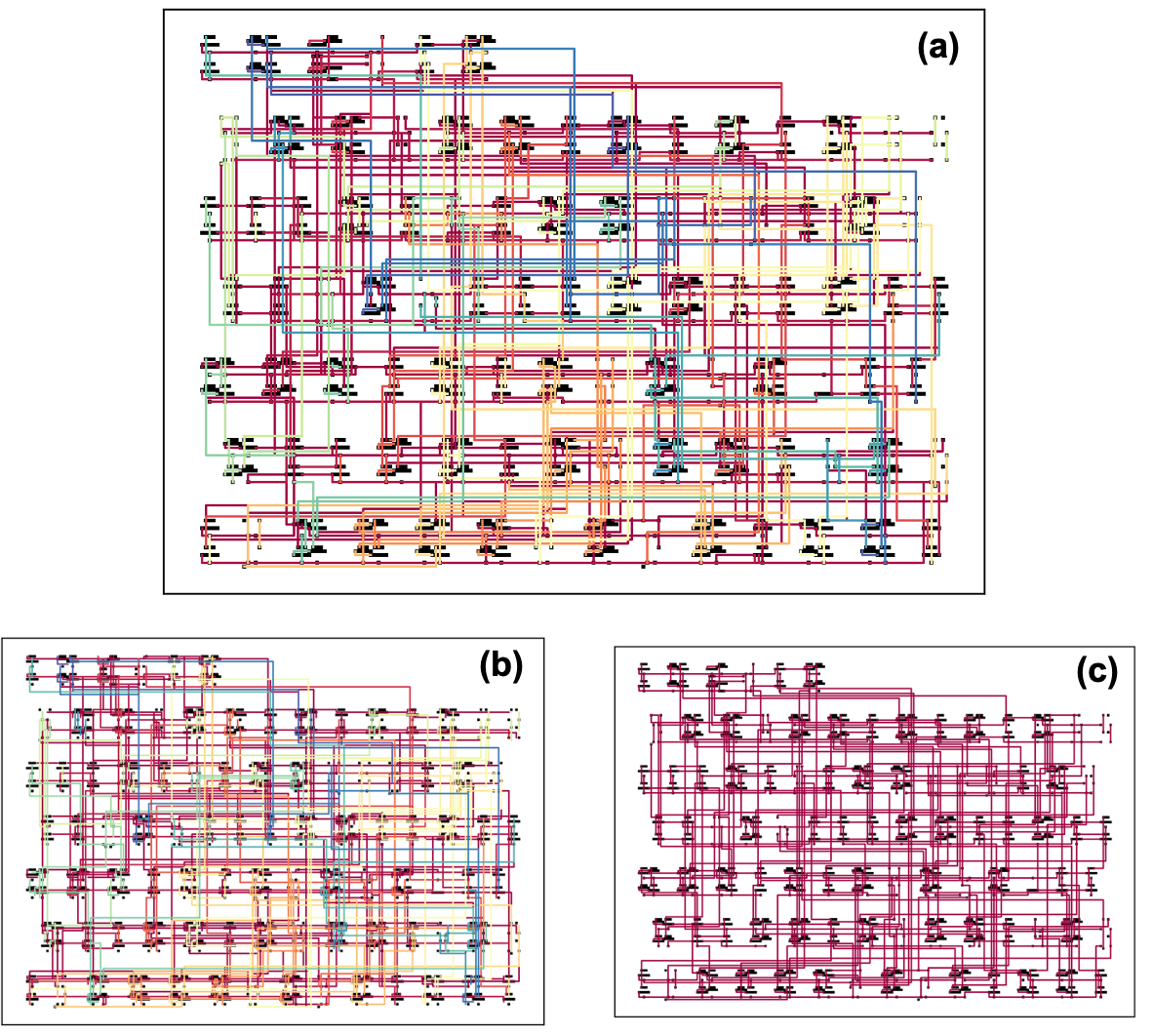}
\caption{Visualization of detailed routing solutions for a \textit{sarfsm500} problem in test data set given by (a) supervised learning policy model, (b) reinforcement learning policy model and (c) genetic algorithm.}
\label{sar_vis}
\end{figure}

Correlation analysis  between SRL and ARL  models on the training and test sets of \textit{biasamp5000} and \textit{sarfsm500} is also conducted. On both data sets, the two models solutions' cost show relatively strong correlations on the training and test, and the $R^2$ scores on the training and test for the same design is similar. However, the $R^2$ scores on the training ($R^2=0.940$) and test set ($R^2=0.935$)  of \textit{biasamp} are higher than those of \textit{sarfsm} (training: $R^2=0.821$,test: $R^2=0.860$). This indicates that the possibility of higher similarity between solutions of the ARL and SRL  models on  \textit{biasamp} than that of \textit{sarfsm}, which can be confirmed by comparing the actual routing solutions of three methods (SRL, ARL and GA) on test set of \textit{sarfsm500} test set (Fig.~\ref{sar_vis}). 

For  \textit{sarfsm}, less similar patterns can be observed between SRL (Fig.~\ref{sar_vis}a) and ARL (Fig.~\ref{sar_vis}b) model's routes; while more similar patterns can also be found between the SRL policy model (Fig.~\ref{sar_vis}a) and  GA (Fig.~\ref{sar_vis}c), than the comparison between the ARL policy model (Fig.~\ref{sar_vis}b) and the GA (Fig.~\ref{sar_vis}c). This shows the possibility that the SRL policy model in better imitating the behavior of methods used for generate supervision. Yet, further statistical analysis on larger data sets is needed to consolidate this observation. Another interesting observation is that, given the divergence behavior of ARL and SRL policy model on \textit{sarfsm} data set, there is still a high correlation between their solutions cost . This shows the possibility that ARL policy model can achieve as good performance of SRL policy model, yet with less similar routing patterns of GA-based routes. The ARL policy model results' similarity to GA is also analyzed: in both training and test sets of \textit{biasamp5000} and \textit{sarfsm500}, the ARL policy model's cost deviation from GA is similar to the pattern displayed by SRL policy model (Fig.~\ref{hist}) with maximum deviation less than $10\%$. In general, the ARL policy model is as good as the SRL policy model given the data sets used in this research.

\section{CONCLUSION}
We present a new supervised learning approach for training the policy model in solving track-assignment detailed routing. The supervision is based on minimizing the KL divergence loss between the output from the policy model and a given distribution from the GA solver. It achieves performance comparable to reinforcement learning trained policy model, indicating the supervised learning as an alternative approach for training the policy model, that can leverage existing expert solutions to obtain user customized routers. Both supervised trained and reinforcement learned policy models have around 100$\times$ acceleration compared with the baseline GA solver. To better understand the policy models' behavior in solving the  sequencing step in detailed routing, a statistical analysis of the results are applied. The results show high correlations  between GA results' cost and policy model results' cost, indicating that policy models can be deployed as a fast and high-quality surrogate model for GA in solving sequential detailed routing. This would enable better routability-driven placement driven optimization: with much less routability evaluation time for a placement solutions, the optimization resources for placement optimization has a two scale increase. By comparing the behavior and working principle of supervised learning and reinforcement learning in training policy models in this work, it is concluded that SRL model is able to behave as well as ARL model, while provides a feasible way for learning from past designs or human experts' designs. Future works including more systematic investigation of sample efficiency of the SRL and ARL policy models. After this, integrating the policy models into placement optimization would also be implemented.






\section{Acknowledgments}
This work is partially funded by the DARPA IDEA program (HR0011-18-3-0010).


\bibliographystyle{ACM-Reference-Format}
\bibliography{ref}

\appendix

\end{document}